\documentclass[lettersize,journal]{IEEEtran}
\usepackage{amsmath,amsfonts}
\usepackage{algorithmic}
\usepackage{algorithm}
\usepackage{array}
\usepackage[caption=false,font=normalsize,labelfont=sf,textfont=sf]{subfig}
\usepackage{textcomp}
\usepackage{stfloats}
\usepackage{url}
\usepackage{verbatim}
\usepackage{booktabs}
\usepackage{graphicx}
\usepackage{cite}
\usepackage{multirow}
\usepackage{hyperref}
\usepackage{threeparttable}
\usepackage{color}
\usepackage{mathtools}
\usepackage{tabularx}
\usepackage{rotating}
\usepackage{pdfpages}

\definecolor{dg}{RGB}{40, 180, 0}

\begin{document}

\title{LVBA: LiDAR-Visual Bundle Adjustment for RGB Point Cloud Mapping}

\author{Rundong Li, Xiyuan Liu, Haotian Li, Zheng Liu, Jiarong Lin, Yixi Cai, and Fu Zhang,~\IEEEmembership{Member,~IEEE,}
    % <-this % stops a space
    \thanks{The authors are with the Department of Mechanical Engineering, The University of Hong Kong, Hong Kong, China (e-mail: \{rdli10010, xliuaa, haotianl, u3007335, zivlin, yixicai\}@connect.hku.hk; fuzhang@hku.hk)}
}

\markboth{Journal of \LaTeX\ Class Files,~Vol.~14, No.~8, August~2021}%
{Shell \MakeLowercase{\textit{et al.}}: A Sample Article Using IEEEtran.cls for IEEE Journals}
% Remember, if you use this, you must call \IEEEpubidadjcol in the second

\maketitle
\pagestyle{empty}
\thispagestyle{empty}
\begin{abstract}
Point cloud maps with accurate color are crucial in robotics and mapping applications. Existing approaches for producing RGB-colorized maps are primarily based on real-time localization using filter-based estimation or sliding window optimization, which may lack accuracy and global consistency. In this work, we introduce a novel global LiDAR-Visual bundle adjustment (BA) named LVBA to improve the quality of RGB point cloud mapping beyond existing baselines. LVBA first optimizes LiDAR poses via a global LiDAR BA, followed by a photometric visual BA incorporating planar features from the LiDAR point cloud for camera pose optimization. Additionally, to address the challenge of map point occlusions in constructing optimization problems, we implement a novel LiDAR-assisted global visibility algorithm in LVBA. To evaluate the effectiveness of LVBA, we conducted extensive experiments by comparing its mapping quality against existing state-of-the-art baselines (i.e., R$^3$LIVE and FAST-LIVO). Our results prove that LVBA can proficiently reconstruct high-fidelity, accurate RGB point cloud maps, outperforming its counterparts.
\end{abstract}


\section{Introduction and Related Works}
\IEEEPARstart{H}{igh}-fidelity colorized 3D maps are vital in diverse fields including robotics\cite{r3live}, environmental science\cite{pavlis2017new}, and civil engineering\cite{fassi2013comparison}. They serve various purposes, which can provide essential data for robot navigation, offer accurate models for environmental monitoring, and form the basis of digital twin simulations.

Based on the used sensors, existing colorized 3D map reconstruction methods can be primarily categorized into three classes: 1) Visual camera-based\cite{pba1, densemonomapping}; 2) RGB-D sensor-based\cite{stereolio, dense_rgbd}; 3) LiDAR-camera fusion based \cite{r3live, fast-livo}. For visual camera-based methods \cite{pba1, densemonomapping}, without direct depth measurements, they triangulate the depth from multi-view geometry. This process usually involves feature tracking algorithms, which can introduce systematic errors and increase the risk of incorrect 3D geometry estimation\cite{pba2}. Methods using RGB-D sensors can utilize direct dense depth measurements through structure light. However, most of them are range-limited due to the intensity constraints of the detection light. In contrast, 3D points measured from LiDAR sensors are accurate, offering a long detection range and errors within centimeters. Hence, reconstructing colorized 3D maps from LiDAR-camera fusion based methods are drawing increasing attention in the literature. In this work, we aim to develop a LiDAR-visual bundle adjustment method to improve the mapping quality of LiDAR-camera fusion-based methods.

\begin{figure}[!pt] 
    \setlength\abovecaptionskip{-0.1\baselineskip}
    \centering
    \includegraphics[width=1.0\linewidth]{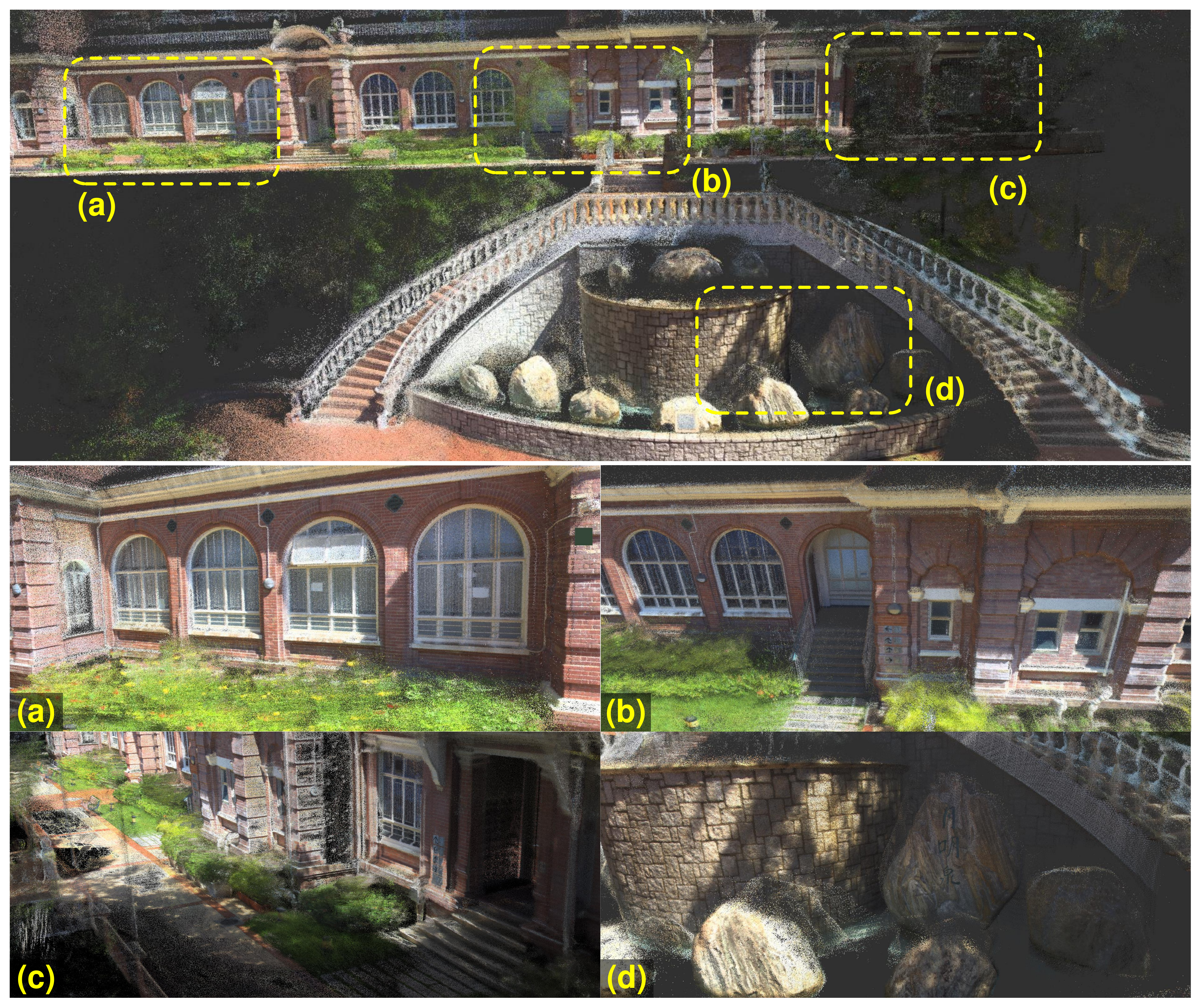}
    \caption{An RGB point cloud map optimized using our method. The depicted data was captured at the Chong Yuet Ming Physics Building at The University of Hong Kong. Our method effectively optimizes both LiDAR and camera poses, achieving high levels of accuracy and consistency in the mapping process.}
    \label{cover_fig}
    \vspace{-0.6cm}
\end{figure}

Although significant progress has been made in many recent approaches (e.g., \cite{laservisualinertial, limo, stereolio, licfusion}), the fusion of LiDAR and camera measurements, as well as the reconstruction of the high-fidelity colorized map, remain challenging. Previous works (e.g., \cite{r3live, fast-livo}) for the LiDAR-visual platform have primarily focused on improving localization accuracy or enhancing the system robustness against sensor degeneration. However, most of these works rely on filter or sliding window methods, which cannot avoid accumulation errors and are susceptible to inaccurate extrinsic parameters. These issues directly affect the observation consistency of camera images, resulting in blurred colorized point clouds.

In fact, these problems can be solved by employing a global optimization method (i.e., bundle adjustment). By directly optimizing the camera's observation consistency using existing geometry information from LiDAR point clouds, globally consistent high-fidelity point clouds can be constructed even without accurate extrinsic parameters.

To our knowledge, the only global optimization method for LiDAR-visual sensor fusion is Colmap-PCD\cite{colmap-pcd}, which is only a recent proposal. Based on the Colmap\cite{schoenberger2016mvs, schoenberger2016sfm} framework, Colmap-PCD registers camera frames to a given LiDAR point cloud using a geometric visual bundle adjustment algorithm.
Compared to Colmap-PCD, a key difference in our method is the use of photometric visual bundle adjustment, which bypasses the need for feature extraction and matching algorithms and shows better robustness under low-texture environments.  Moreover, by effectively utilizing LiDAR scan information, we address the occlusion check without ray-tracing and effectively build up the global constraint. This approach enhances the optimization's efficiency, as the ray-tracing process is typically time-consuming.
Another advantage of photometric visual bundle adjustment is its ability to estimate camera exposure parameters (e.g., exposure time, gain, etc.)\cite{dso}, which could directly affect the image brightness and is always crucial. The existing works often directly estimate a scale factor called "exposure time"\cite{dso, r3live++} without considering the noise model and covariance, which can lead to brightness drift during global bundle adjustment. In this paper, we derive a relative exposure time with covariance consideration, allowing us to estimate exposure time globally.

In summary, we propose a LiDAR-visual bundle adjustment (LVBA) to optimize camera and LiDAR poses. LVBA works in two stages: first, it optimizes LiDAR poses through a LiDAR BA\cite{balm2}; then, it optimizes camera poses using a photometric BA method, leveraging the geometric prior provided by the LiDAR point cloud. Our main contributions are:
\begin{itemize}
    \item We proposed a photometric bundle adjustment to estimate camera states using the prior LiDAR point cloud map, which improves the colorization quality of the point cloud even without accurate time alignment or finely calibrated extrinsic parameters.
    \item We proposed a LiDAR-assisted scene point generation and visibility determination algorithm, which contains global co-visibility of camera frames that facilitates us to construct the global photometric visual bundle adjustment problem.
    \item We implemented a toolchain to evaluate the accuracy and consistency of the colorized maps.
    Using this toolchain, we conducted an extensive evaluation of our proposed LVBA against state-of-the-art LiDAR-Visual-Inertial mapping approaches. Our evaluation results demonstrate that LVBA outperforms other state-of-the-art works in accurately and consistently reconstructing colorized point cloud maps (refer to Fig. \ref{cover_fig}).
\end{itemize}
\section{Methodology}
\subsection{System Overview}
\begin{figure}[!t] 
    \centering
    \includegraphics[width=1.0\linewidth]{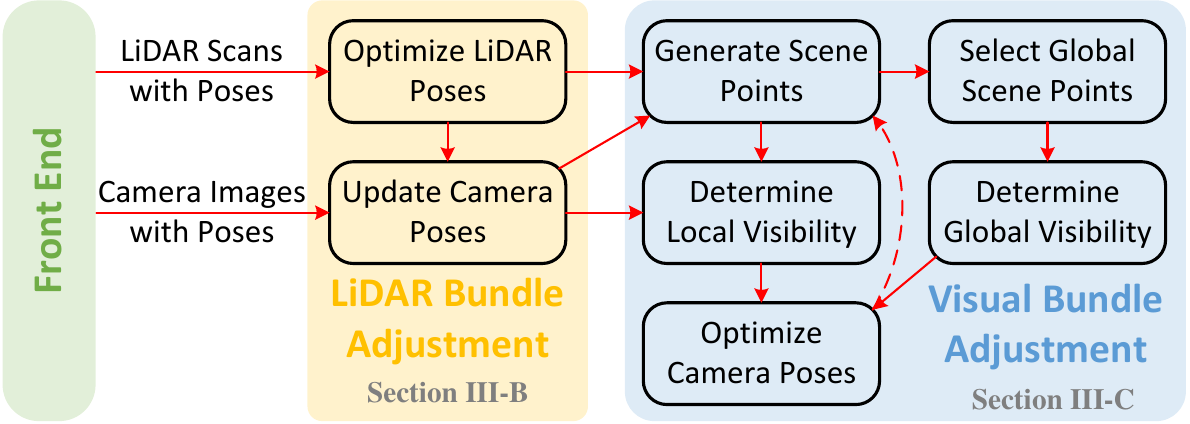}
    \caption{The overview of our system. Our system consists of a LiDAR BA and a visual BA.}
    \label{system_overview}
    \vspace{-0.5cm}
\end{figure}

The overview of our proposed LiDAR-visual bundle adjustment (LVBA) is shown in Fig. \ref{system_overview}. LVBA takes LiDAR scans, camera images, and their rough poses within the same world reference frame as input, typically from a front-end method such as \cite{r3live, fast-livo}. The system includes two components, namely the LiDAR BA part and the visual BA part. Firstly, the LiDAR states undergo optimization using a LiDAR BA method\cite{balm2}, producing an optimal LiDAR pose estimation and planar features. Subsequently, the camera poses are optimized through an iterative coarse-to-fine photometric bundle adjustment, which utilizes LiDAR scans and extracted planar feature points as prior.
\subsection{LiDAR Bundle Adjustment}
In LVBA, we utilize a LiDAR BA method called BALM\cite{balm, balm2} (version 2) to optimize the LiDAR poses. This method formulates the LiDAR BA problem by leveraging the edge and plane features extracted from LiDAR point clouds. The optimization process seeks to minimize the Euclidean distance between each point in a scan and its neighboring edge or plane. We use BALM to optimize the 6 DoF pose of each LiDAR scan and to build a voxel map that contains plane features extracted from the LiDAR points, which is then used in our later process of visual BA. Denote $\widehat{\mathbf T}_L$ the LiDAR pose after BALM optimization.

To ensure alignment between the camera trajectory and optimized LiDAR poses, we also need to update the camera poses after BALM optimizes the LiDAR pose. To do so, for each camera frame $\mathbf T_C\in SE(3)$ before BALM optimization, the closest LiDAR frame $\mathbf T_L\in SE(3)$ before BALM optimization is found. Note that $\mathbf T_C$ and $\mathbf T_L$ are respectively the camera and LiDAR poses in the same world reference frame. Then the camera pose $\mathbf T_C$ should be updated as:
\begin{align}
    \widehat{\mathbf T}_C = \widehat{\mathbf T}_L \mathbf T_{L}^{-1} \mathbf T_{C}
\end{align}
\subsection{Visual Bundle Adjustment with LiDAR Prior}
The visual BA in LVBA first generates visual feature points named ``scene points'' and then formulates the cost function by projecting scene points onto different image frames and minimizing the photometric discrepancy between them. Two types of scene points are used, local scene points and global scene points. Sec. \ref{sec: scene_point_generation} and \ref{sec: local_visibility} present local scene points selection and visibility determination. Sec. \ref{sec: global_visibility} presents global scene point selection and visibility determination. Finally, Sec. \ref{sec: photometric_error_formulation} and \ref{sec: lm_optimization} describe the process of constructing our photometric cost and its optimization.

\subsubsection{Local Scene Point Generation}
\label{sec: scene_point_generation}
Our visual BA begins by generating local scene points for each image frame. When selecting local scene points, we prefer points surrounded by complex textures as they provide richer photometric details and effective constraints when projected onto an image frame. Specifically, for each camera frame, its image is divided into grid cells. The LiDAR planar feature points captured when the position of LiDAR is close to this camera frame are then projected onto these grid cells.
To ensure the view quality of the scene point, only those LiDAR feature points whose surface normal is alongside the view direction from the camera frame are taken into consideration. Specifically, the projected LiDAR feature points should fulfill the following condition:
\begin{align}
    \label{eq. sp_gen}
    \Bigg| \mathbf n_f^T \frac{(\mathbf p_f- \mathbf t_C)}{\|\mathbf p_f-\mathbf t_C\|_2} \Bigg| > \alpha_0
\end{align}
where $\mathbf p_f$ and $\mathbf n_f$ are the position and normal vector of the LiDAR planner feature point, respectively. $\mathbf t_C$ is the position of the camera frame, and $\alpha_0$ is a threshold.
After each LiDAR feature point is projected, a score reflecting the intensity gradients around the candidate scene point is subsequently computed for every projected point by employing the Difference of Gaussian (DoG) method \cite{marr1980theory}. Within each grid cell, the point with the highest score is then selected as a local scene point. These selected local scene points are denoted as $\boldsymbol \pi=(\mathbf p, \mathbf n)\in \mathbb R^3\times \mathbb S^2$, where $\mathbf p$ represents the position of the point and $\mathbf n$ represents the normal vector estimated by our LiDAR BA. For each selected local scene point, we denote the corresponding camera frame as the ``reference frame" of this local scene point. In scenarios where a grid cell only contains points with low scores (indicating a lack of useful photometric information, such as in a textureless surface), no points are selected from that grid cell.
\subsubsection{Local Visibility Determination}
\label{sec: local_visibility}
After generating local scene points for each frame, our next step is to identify the other frames that can observe these points. These identified frames are then denoted as the ``target frame'' of each scene point and will be then used to construct the corresponding cost item in (\ref{eq:cost_item}). To facilitate this, a local visibility determination process is implemented. In the process, only frames within a sliding window relative to the reference frame of a local scene point are considered. Since these frames are close to the position of the reference frame, hence the parallax is insignificant and a local scene point is usually directly visible by these frames without further occlusion check. As a consequence, we only have to examine if the view direction between the local scene point $\boldsymbol \pi=(\mathbf p, \mathbf n)$ and the candidate target frame at $\mathbf T_t=(\mathbf R_t, \mathbf t_t)$ is within the frame's FoV. The FoV check is much more efficient than the occlusion check based on ray-casting but may give false visibility results for scene points on the edge of foreground and background objects. To fix this issue, we project, using the initial sensor pose, the local scene point onto the reference frame, and the candidate target frame to evaluate the photometric discrepancy between them. If a scene point is truly visible in the target frame, the initial photometric discrepancies are often small.  Finally, to ensure the view quality of the scene point, we choose target frames that are viewing the scene point along a direction close to the point surface normal. %also check the angle between the normal vector $\mathbf n$ of the local scene point and view direction, if the view direction is almost vertical to the normal vector of the local scene point, then the point cannot be well observed. 
Consequently, the local visibility determination checks are:
\begin{align}
\label{eq.vis_determination}
    | \mathbf d^T \mathbf z_C | > \alpha_1, \quad | \mathbf d^T \mathbf n | > \alpha_2, \quad \mathtt{NCC}(\mathbf I_r, \mathbf I_t) > \alpha_3
\end{align}
where $\mathbf d=\frac{\mathbf p-\mathbf t_t}{\|\mathbf p-\mathbf t_t\|_2}$ denotes the view direction from the candidate target frame to the local scene point, $\mathbf z_C$ express the z-axis of the camera frame (vertical to the camera image plane), $\mathbf I_r$ and $\mathbf I_t$ are the RGB values of patches generated on the reference frame and the candidate target frame, respectively, and $\mathtt{NCC}(\cdot, \cdot)$ is the Normalized Cross-Correlation (NCC) \cite{brunelli2009template} between the two patches. The details for patch generation are illustrated in Sec. \ref{sec: photometric_error_formulation}. $\alpha_1$, $\alpha_2$, and $\alpha_3$ are three thresholds. If a local scene point is deemed visible by this frame (i.e., satisfying (\ref{eq.vis_determination})), it will be selected as a target frame and contribute to the final optimization cost in (\ref{eq:cost_item}).
\subsubsection{Global Scene Point Selection and Visibility Generation}
\label{sec: global_visibility}

\begin{figure}[!t]
    \setlength\abovecaptionskip{-0.6\baselineskip}
    \centering
    \includegraphics[width=0.9\linewidth]{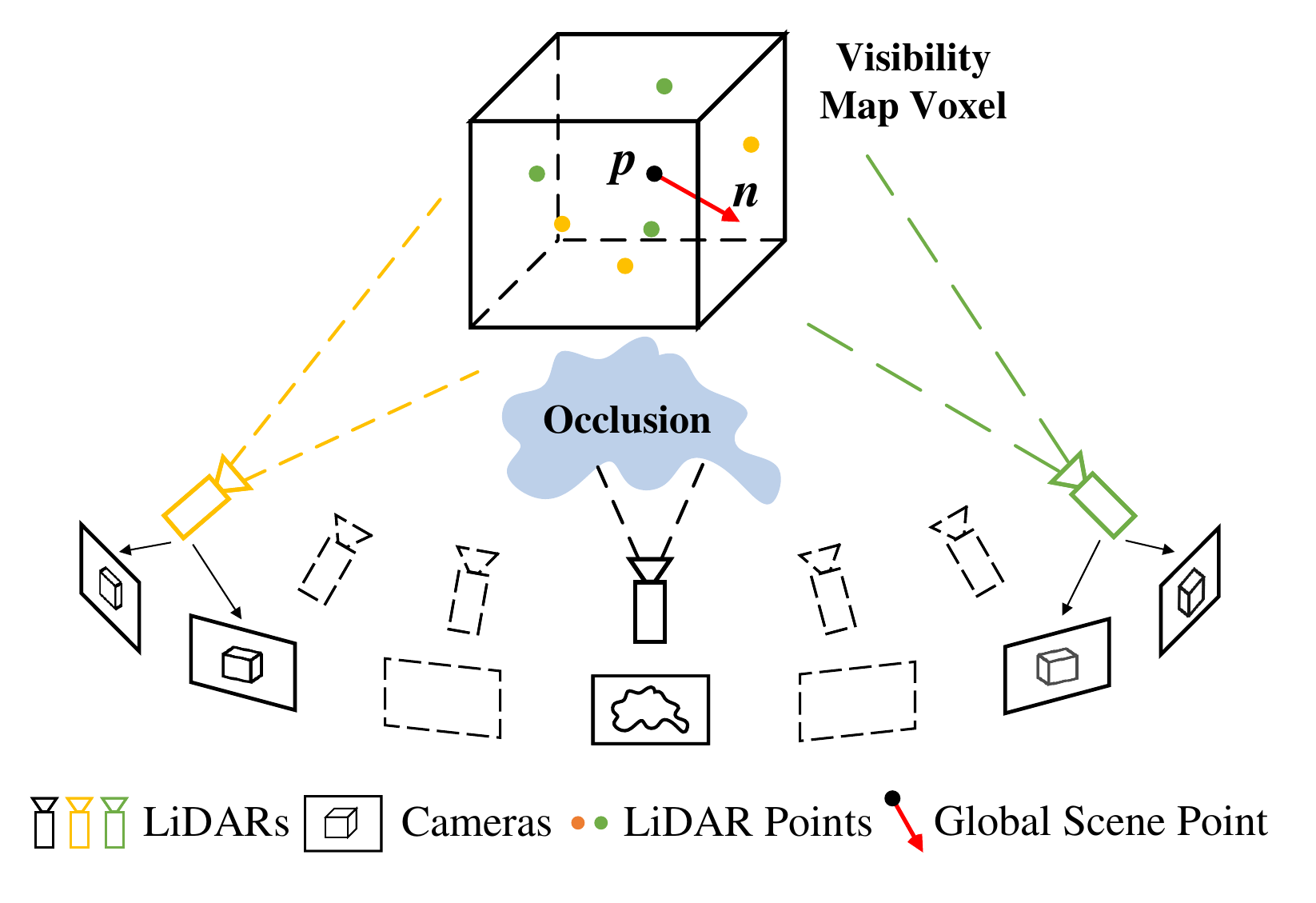}
    \caption{LiDAR-assisted Global Visibility Map. A voxel map that stores the viability information of each voxel is constructed with LiDAR scans (\ref{sec:global_visibility_map}). After that, a global scene point is selected from all scene points in each voxel (Sec. \ref{sec: global scene point selection}). The visibility voxel map, together with the selected global scene point is then used in the global visibility determination process (Sec. \ref{sec: Global Visibility Determination})}
    \label{fig:global_visibility}
    \vspace{-0.5cm}
\end{figure}

Local scene points provide effective local constraints for camera pose optimization. However, they fail to provide more global constraints for camera poses far apart observing the same area.  Therefore, we introduce global scene points and global visibility that provide more comprehensive, global constraints. When generating global scene points and determining global visibility, the distance between the reference and target frames increases, making the occlusion check necessary. Typically, a ray-casting algorithm (e.g., in \cite{kong2023marsim}) is employed to address the occlusion check, but this approach not only increases the computation costs but also risks inaccuracies in sparser point cloud scenarios. To overcome these challenges, we implement a LiDAR-assisted method based on a global visibility voxel map, addressing the occlusion check with more reliable results and reduced computational load, as illustrated in Fig. \ref{fig:global_visibility}.
\paragraph{Global Visibility Voxel Map}
\label{sec:global_visibility_map}
Our method for global scene point selection and visibility determination relies on a global visibility voxel map created using LiDAR scan data. This map records the visibility information for each voxel and is constructed after the optimization of LiDAR poses using BALM. During construction, for each point in a LiDAR scan,  the voxel it belongs to is appended with the pose of the LiDAR scan, indicating that the voxel is visible from the position of this LiDAR scan. Subsequently, camera frames proximate to these visible positions are identified and stored. Consequently, each voxel in the map retains a set of indices of camera frames from which it is visible.

\paragraph{Global Scene Point Selection}
\label{sec: global scene point selection}
To limit the computation cost, only a subset of all local scene points, namely global scene points, is used to construct global constraints. To select these global scene points, we distribute all local scene points into the visibility voxel map. For each voxel, the local scene point that exhibits the best observation quality from their respective reference frames is selected. To model the observation quality, we define a score $s$, which describes the projection area on the reference image of a small surface element at the scene point. A larger projection area indicates better observation quality. More specifically, for each candidate scene point $\boldsymbol \pi=(\mathbf p, \mathbf n)$ the score $s$ is calculated by:
\begin{align}
    s=\frac{\mathbf n^T (\mathbf p-\mathbf t_r)}{\| \mathbf p-\mathbf t_r \|^2_2}
\end{align}
where $\mathbf t_r$ is the position of the corresponding reference frame. The local scene point within each voxel that achieves the highest score is then selected as a global scene point.

\paragraph{Global Visibility Determination}
\label{sec: Global Visibility Determination}
Once the global scene points have been selected, the next step is ascertaining their visibility across all camera frames. This process comprises two main steps: First, we identify, utilizing the global visibility map, the camera frames stored in the voxel where the global scene point belongs as the candidate target frames. This is because these camera frames are most likely to observe this voxel, and hence the global scene point in it. Second, we identify those candidate target frames satisfying  (\ref{eq.vis_determination}) as the true target frames, following the same reason as in local visibility determination.
\subsubsection{Photometric Error Formulation}
\begin{figure}[!t]
    \setlength\abovecaptionskip{-0.1\baselineskip}
    \centering
    \includegraphics[width=0.9\linewidth]{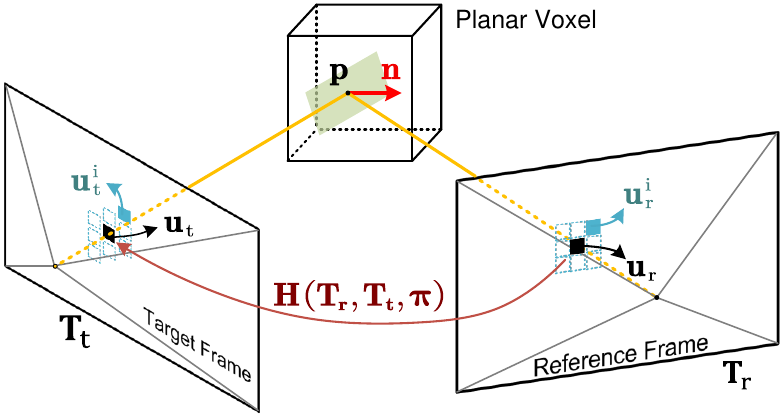}
    \caption{Photometric Error Formulation: A scene point $\boldsymbol \pi=(\mathbf p, \mathbf n)$ is first projected to the reference image frame at $\mathbf T_r$, and a reference patch $\{\mathbf u_r^{(i)}\}$ is generated. Then, the reference patch is projected and wrapped onto the target image frame at $\mathbf T_t$ by a homography transformation $\mathbf H$ to generate a target patch $\{\mathbf u_t^{(i)}\}$. Finally, a photometric error is constructed with the $L2$-norm of the radiance error between two patches.}
    \label{fig: photometric_error_formulation}
    \vspace{-0.5cm}
\end{figure}
\label{sec: photometric_error_formulation}
The visual BA in LVBA optimizes the photometric discrepancy of a scene point $\boldsymbol \pi=(\mathbf p,\mathbf n)\in \mathbb{R}^3\times \mathbb{S}^2$  (either local or global scene points) between the reference frame and target frame, whose poses are denoted as $\mathbf T_r=(\mathbf R_r, \mathbf t_r)\in SE(3)$ and $\mathbf T_t=(\mathbf R_t, \mathbf t_t)\in SE(3)$, respectively. As shown in Fig. \ref{fig: photometric_error_formulation}, a cost item is constructed below:

Firstly, the scene point $\boldsymbol \pi$ is projected to the reference image plane via the pin-hole camera projection model:
\begin{align}
    \hat{\mathbf u}_r=\mathbf{K} \mathbf R_r^T(\mathbf p-\mathbf t_r)
\end{align}
where $\mathbf K$ is the camera intrinsic matrix, $\mathbf {\hat u}_r\in \mathbb R^3$ is the homogeneous format of $\mathbf{u}_r\in \mathbb R^2$ in the reference image plane coordinate (without further notice, we use $\hat{(\cdot)}$ to denote the homogeneous format of a vector). Then, a reference patch $\{\mathbf u_r^{(i)}\}$ is generated around the projected point.
Utilizing the plane normal of the scene point and the camera projection model, the reference patch can be projected to the target frame via a homography transformation\cite{mvgeo}:
\begin{align}
    \hat{\mathbf u}_t^{(i)} &= \mathbf H(\mathbf T_r, \mathbf T_t, \boldsymbol \pi) \hat{\mathbf u}_r^{(i)}
\end{align}
where the homography matrix $\mathbf H$ can be expressed as
\begin{align}
    \mathbf H &= \mathbf K\mathbf R_t^{-1}\big[ \mathbf n^T(\mathbf p-\mathbf t_r)\mathbf E+(\mathbf t_r-\mathbf t_t)\mathbf n^T \big]\mathbf R_r\mathbf K^{-1}
\end{align}
and $\mathbf E$ is $3\times 3$ identity matrix. Finally, the photometric cost is constructed as:
\begin{align}
    \label{eq:cost_item}
    L_{\text{photo}}(\boldsymbol{\mathcal T}_r, \boldsymbol{\mathcal T}_t; \boldsymbol \pi) &= \sum_{i=0}^{N^2}\|\epsilon_t^{-1} \mathbf I_t(\mathbf u_t^{(i)}) - \epsilon_r^{-1} \mathbf I_r(\mathbf u_r^{(i)})\|_{\boldsymbol \Sigma}^2 \; \nonumber \\
    \boldsymbol \Sigma &= \bigg(\frac{1}{\epsilon_r^2+\epsilon_t^2}\bigg)\mathbf E
\end{align}
where $\mathbf I_r(\boldsymbol \cdot)$ and $\mathbf I_t(\boldsymbol \cdot)$ calculate RGB color vector by bi-linear interpolation on the reference image and target image, respectively, $\boldsymbol{\mathcal T}_r=(\mathbf T_r, \epsilon_r)$ and $\boldsymbol{\mathcal T}_t=(\mathbf T_t, \epsilon_t)$ represent the camera states of reference and target frames, $\epsilon_r$ and $\epsilon_t$ represent the relative exposure time of them and $\boldsymbol \Sigma$ is the covariance matrix of the cost function. The relative exposure time $\epsilon$ and covariance matrix $\boldsymbol \Sigma$ are defined using a simplified camera measurement model. Further details can be referred to the APPENDIX A\footnote{\url{https://github.com/hku-mars/mapping_eval}}.

\subsubsection{Levenberg-Marquardt Optimization}
\label{sec: lm_optimization}
With our constructed photometric cost item, the optimal estimation of camera states $\boldsymbol{\mathcal T}^\ast$ is given by:
\begin{align}
\label{eq: cost function}
    \boldsymbol{\mathcal T}^\ast &= \underset{\boldsymbol{\mathcal T}}{\operatorname{argmin}}{\sum_{i=1}^{M_s}\sum_{j\in V_i}{L_{\text{photo}}(\boldsymbol{\mathcal T}_{r(i)}, \boldsymbol{\mathcal T}_j; \boldsymbol \pi_i)}}
\end{align}
where $\boldsymbol{\mathcal T} = \{\boldsymbol{\mathcal T}_1, \boldsymbol{\mathcal T}_2, \cdots, \boldsymbol{\mathcal T}_{M_p}\}$ is the camera state set, $\boldsymbol \Pi = \{\boldsymbol \pi_1, \boldsymbol \pi_2, \cdots, \boldsymbol \pi_{M_s}\}$ is the scene point set (both local and global), $M_p$ and $M_s$ represents the total number of camera poses and scene points, respectively, $r(i)$ represents the reference frame index of the $i$-th scene point, $V_i$ is the target frame set (both local and global), and $L_{\text{photo}}$ is the photometric cost item, as defined in (\ref{eq:cost_item}). In LVBA, we employed a Levenberg-Marquardt optimizer to minimize the cost function in (\ref{eq: cost function}).

To enhance the robustness of the cost function against initial estimation, our visual BA employs an iterative coarse-to-fine strategy. This strategy involves the process of down-sampling the image and sequentially optimizing the camera states from the top layer of the pyramid to the original resolution. Within each iteration, we utilize the optimized states obtained from the previous higher layer to generate scene points and determine visibility.

\section{Experiments}
\label{sec.exp}
\subsection{Experiment Setup}
{During the experiment, we compared our LVBA with other state-of-the-art LiDAR-visual(-inertial) sensor fusion methods, including R$^3$LIVE\cite{r3live}, FAST-LIVO\cite{fast-livo}, and Colmap-PCD\cite{colmap-pcd}. We perform evaluations on three datasets, the R$^3$LIVE dataset\cite{r3live}, the FAST-LIVO dataset\cite{fast-livo}, and MaRS-LVIG\cite{mars-datset}. We conducted mapping accuracy evaluation on all three datasets, and the trajectory accuracy evaluation on MARS-LVIG, which provided ground truth trajectory.}

For both the Colmap-PCD and LVBA, we utilize R$^3$LIVE to provide the initial estimation when evaluating on the R$^3$LIVE dataset and MaRS-LVIG dataset, while employing FAST-LIVO to provide the initial estimation when evaluating on the FAST-LIVO dataset. Since these data are collected at a high rate, typically, 10 Hz, optimizing all frame poses on the raw data is computationally demanding for LVBA and Colmap-PCD, which are two global optimization approaches.
To constrain the computation load, we extracted keyframes for camera frames.
Furthermore, since some of the sequences are extensively long for Colmap-PCD and LVBA, leading to excessive time and memory costs, we separated them into sub-sequences, and the average value of the evaluation result for each sequence is then taken as the final result.
\subsection{Mapping Evaluation}
\subsubsection{Evaluation Method}
\begin{figure}
    \setlength\abovecaptionskip{-0.1\baselineskip}
    \centering
    \includegraphics[width=0.9\linewidth]{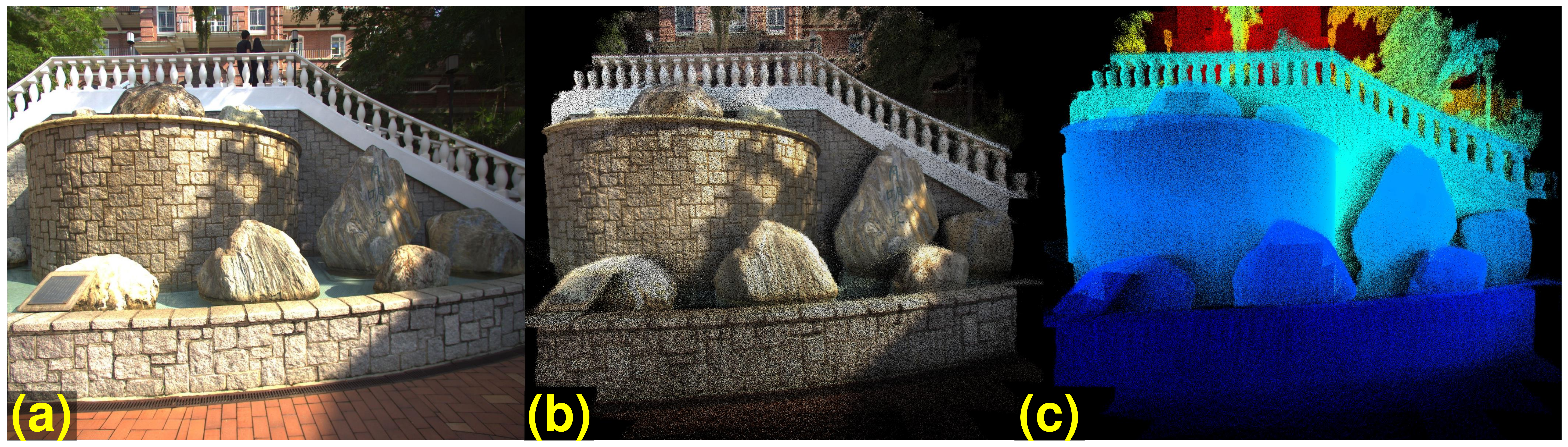}
    \caption{A sample output from our mapping evaluation algorithm. (a) displays the original image captured by the camera. (b) shows the image as rendered by our algorithm, and (c) presents the depth map, derived from the point cloud, which was used in the rendering of image (b). The rendered image (b) is then compared to the original (a) for evaluation using metrics such as Peak Signal-to-Noise Ratio (PSNR) and Structural Similarity Index Measure (SSIM).}
    \label{fig:eval_result}
    \vspace{-0.5cm}
\end{figure}
To evaluate the mapping accuracy and consistency with our optimization results, we use a colorizing-and-rendering algorithm. Initially, an RGB-radiance point cloud is created using LiDAR scans and camera images. For each LiDAR point $\mathbf p^L$ in the LiDAR frame $\mathbf T_L=(\mathbf R_L, \mathbf t_L)$, the nearest camera frame at $\mathbf T_C=(\mathbf R_C,\mathbf t_C)$ is identified. The radiance value of this point is calculated as:
\begin{align}
    \mathbf{r} = \epsilon^{-1} \mathbf I\Big(\boldsymbol \Pi \big(\mathbf R_C^T(\mathbf R_L \mathbf p^L + \mathbf t_L - \mathbf t_C\big)\Big)
\end{align}
where $\epsilon$ is the estimated relative exposure time of the camera frame, for benchmark methods that didn't estimate the relative exposure time, $\epsilon$ is set to $1$, $\boldsymbol \Pi(\cdot)$ denotes the projection process to the camera's image plane, and $\mathbf I(\cdot)$ yields the RGB value. After constructing the radiance map, we render raw images from the radiance map. For each image with optimized camera frame pose and relative exposure time $\epsilon$, we project the colorized radiance map onto its image plane to create a rendered image, the RGB color of the image is determined by $\mathbf I = \epsilon \mathbf r$. An example of our rendering result is shown in Fig. \ref{fig:eval_result}. Finally, the discrepancy between these rendered images and the original raw images is evaluated using Peak Signal-to-Noise Ratio (PSNR) and Structural Similarity Index Measure (SSIM) metrics. The source code of our evaluation toolchain is available on GitHub\footnote{\url{https://github.com/hku-mars/mapping_eval}}.

\subsubsection{Evaluation Setup}
In this section, we performed a mapping accuracy evaluation among all three datasets. Due to the strict time synchronization requirement of FAST-LIVO, which is not satisfied in the R$^3$LIVE dataset, we didn't evaluate FAST-LIVO with the R$^3$LIVE dataset.
Further, as LVBA and Colmap-PCD only estimate the poses of the keyframes, we assessed R$^3$LIVE and FAST-LIVO only at those keyframes, although their results are obtained by running on all frames at 10Hz.

To assess the effectiveness of our proposed global scene point selection and their visibility determination, we ran our algorithm without utilizing any global scene point constraints (termed ``w/o GSP''). Furthermore, to evaluate the effectiveness of our proposed relative exposure time estimation, we ran the algorithm by assuming all relative exposure time to be one (termed ``w/o RET'').
Both experiments were conducted under the same settings of our full method as illustrated above.

\subsubsection{Comparison Results}
\label{sec.mapping_eval}
\newcommand{\thicktoprule}{\toprule[1.5pt]}
\newcommand{\thickbottomrule}{\bottomrule[1.5pt]}
\begin{table*}
\setlength{\tabcolsep}{4.0pt}
\centering
\begin{threeparttable}
\caption{Result of Mapping Evaluation\label{tab:map_eval}}
\begin{tabular}{c|c|c|c|c|c|c|c|c|c|c|c|c|c}
\thicktoprule
\multirow{2}{1cm}{Datasets} & \multirow{2}{*}{Sequences} & \multicolumn{2}{c|}{R$^3$LIVE\cite{r3live}} & \multicolumn{2}{c|}{FAST-LIVO\cite{fast-livo}} & \multicolumn{2}{c|}{Colmap-PCD\cite{colmap-pcd}} & \multicolumn{2}{c|}{Ours (w/o GSP$^1$)} & \multicolumn{2}{c|}{Ours (w/o RET$^2$)} & \multicolumn{2}{c}{\textbf{Ours (Full)}} \\
\cline{3-14}
 && PSNR$\uparrow$ & SSIM$\uparrow$ & PSNR$\uparrow$ & SSIM$\uparrow$ & PSNR$\uparrow$ & SSIM$\uparrow$ & PSNR$\uparrow$ & SSIM$\uparrow$ & PSNR$\uparrow$ & SSIM$\uparrow$ & PSNR$\uparrow$ & SSIM$\uparrow$ \\
\hline
\multirow{8}{1cm}{R$^3$LIVE\cite{r3live}}
&\textit{hku campus seq 00} & 18.13 & 0.1741 & -- & -- & 21.36 & 0.3629 & 20.56 & 0.3328 & 21.06 & 0.3177 & \textbf{21.64} & \textbf{0.3852} \\
&\textit{hku campus seq 01} & 17.82 & 0.1995 & -- & -- & -- & -- & 19.72 & 0.3478 & 19.21 & 0.2905 & \textbf{20.14} & \textbf{0.3750} \\
&\textit{hku campus seq 02} & 16.97 & 0.1520 & -- & -- & -- & -- & 18.91 & 0.2965 & 18.31 & 0.2042 & \textbf{19.34} & \textbf{0.3193} \\
&\textit{hku campus seq 03} & 17.55 & 0.1827 & -- & -- & 17.45 & 0.2267 & 19.16 & 0.3262 & 18.59 & 0.2326 & \textbf{19.56} & \textbf{0.3544} \\
& \textit{hkust campus 00} & 18.03 & 0.1559 & -- & -- & -- & -- & 20.14 & 0.2791 & 19.71 & 0.2361 & \textbf{20.52} & \textbf{0.2934} \\
& \textit{hkust campus 01} & 17.91 & 0.1783 & -- & -- & -- & -- & 20.13 & 0.3006 & 19.64 & 0.2675 & \textbf{20.38} & \textbf{0.3159} \\
& \textit{hkust campus 02} & 16.47 & 0.1628 & -- & -- & -- & -- & 18.23 & 0.2631 & 18.29 & 0.2520 & \textbf{18.90} & \textbf{0.3263} \\
& \textit{hkust campus 03} & 17.54 & 0.1849 & -- & -- & -- & -- & 19.44 & 0.3014 & 19.27 & 0.2548 & \textbf{19.79} & \textbf{0.3285} \\
\hline
\multirow{3}{1cm}{FAST-LIVO\cite{fast-livo}}
& \textit{hku1} & 19.02 & 0.1278 & 21.32 & 0.2052 & 22.45 & 0.3047 & 21.89 & 0.2773 & 22.23 & 0.2810 & \textbf{22.46} & \textbf{0.3364} \\
& \textit{hku2} & 23.69 & 0.2813 & 25.85 & 0.3639 & 26.71 & \textbf{0.4377} & 25.54 & 0.3483 & 26.84 & 0.4212 & \textbf{26.86} & 0.4223 \\
& \textit{visual challenge} & 19.71 & 0.1027 & 21.84 & 0.1378 & 22.83 & 0.1976 & 21.23 & 0.1483 & 22.53 & 0.1786 & \textbf{24.11} & \textbf{0.3377} \\
\hline
\multirow{4}{1cm}{MaRS-LVIG\cite{mars-datset}}
& \textit{HKisland 01} & 15.07 & 0.1061 & 14.50 & 0.0855 & 15.68 & 0.1656 & 17.95 & 0.3466 & 19.34 & 0.4872 & \textbf{19.71} & \textbf{0.5204} \\
& \textit{AMvalley 01} & 22.69 & 0.1716 & 21.16 & 0.1198 & 22.23 & 0.2171 & 26.08 & 0.4386 & 26.44 & 0.4852 & \textbf{27.21} & \textbf{0.5190} \\
& \textit{AMtown 01} & 19.90 & 0.1237 & 19.42 & 0.0854 & 19.17 & 0.1654 & 21.94 & 0.2274 & 22.40 & 0.2631 & \textbf{22.54} & \textbf{0.2679} \\
& \textit{HKairport\_GNSS 01} & 16.77 & 0.0832 & 16.30 & 0.0708 & 18.89 & 0.1799 & 18.10 & 0.1564 & 19.24 & 0.2228 & \textbf{19.54} & \textbf{0.2652} \\
\bottomrule[1.5pt]
\end{tabular}
\begin{tablenotes}
    \footnotesize
    \item $^1$GSP: Global scene point. \quad $^2$ RET: Relative exposure time.\quad $^3$ --\quad: System failed.
\end{tablenotes}
\end{threeparttable}
\vspace{-0.5cm}
\end{table*}
\begin{figure}
    \setlength\abovecaptionskip{-0.2\baselineskip}
    \centering
    \includegraphics[width=0.9\linewidth]{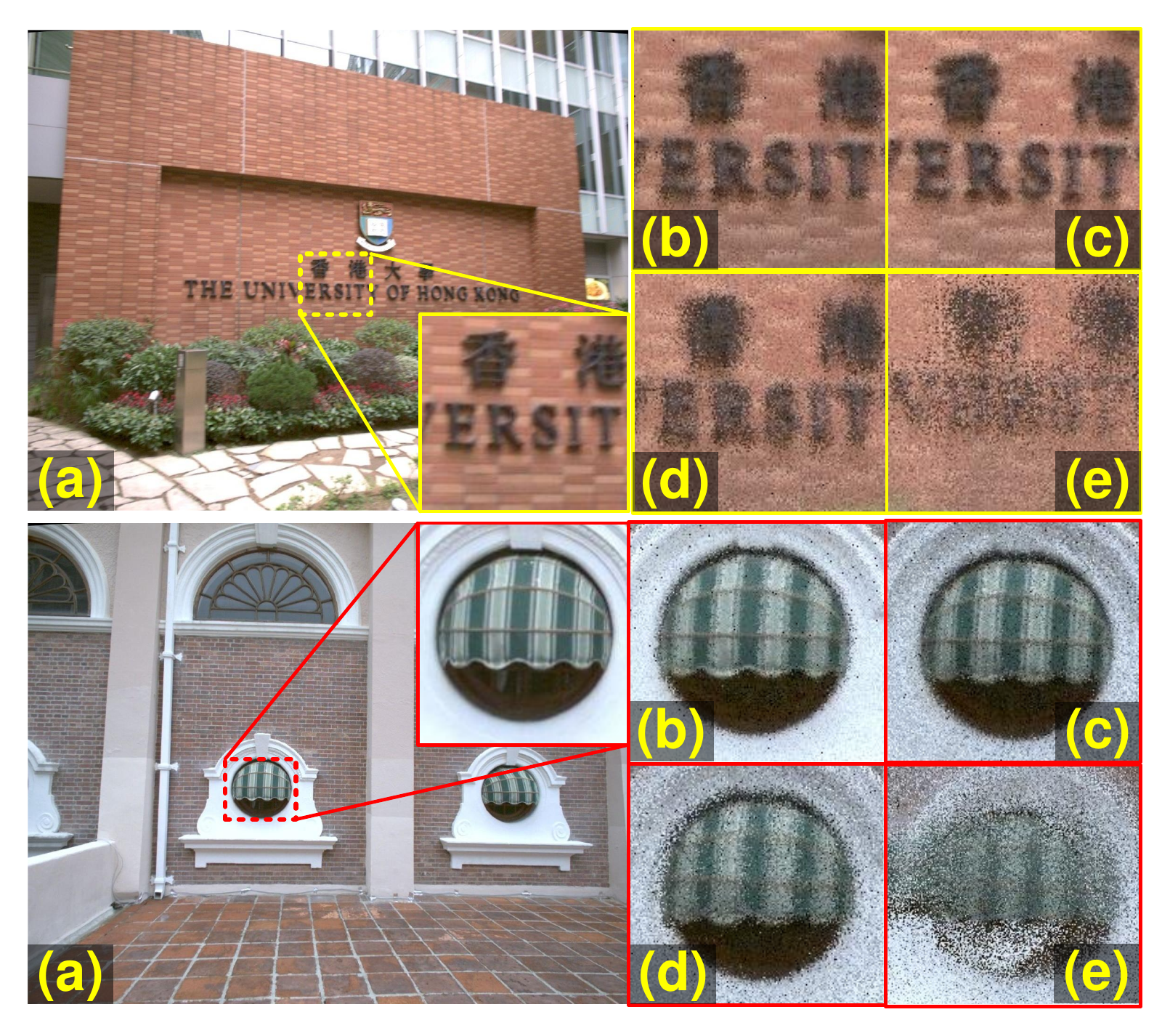}
    \caption{The result of our benchmark experiment. Displayed from left to right: (a) depicts the original image captured by the camera. (b) shows the image generated using our proposed method. (c) shows the image generated using Colmap-PCD\cite{colmap-pcd}. (d) and (e) illustrate the rendered result utilizing state estimations from FAST-LIVO\cite{fast-livo} and R$^3$LIVE\cite{r3live}, respectively.} 
    \label{fig:benchmark}
    \vspace{-0.7cm}
\end{figure}
The evaluation results are computed and reported in TABLE \ref{tab:map_eval}. Compared with two LiDAR-visual-inertial odometry (i.e. R$^3$LIVE and FAST-LIVO), our LVBA (Full) demonstrates significant improvements in mapping accuracy across all tested sequences. Both R$^3$LIVE and FAST-LIVO are based on the ESIKF framework and lack the ability to effectively correct historical errors. When {colorizing} the point cloud with all images, those image frames with substantial state errors are revealed by the color blur of point clouds, especially when the sequences get longer. In contrast, our proposed global LiDAR-visual BA could optimize the state estimation of all image frames, leading to a global consistency and low color blur, as illustrated in Fig. \ref{fig:benchmark}.
\label{sec.colmap-pcd-ana}
Compared with Colmap-PCD, our LVBA (Full) achieved better performance on most of the sequences, except the SSIM on \textit{hku2} in the FAST-LIVO dataset. Since the sequence \textit{hku2} is captured in a night scenario, the image is taken with a high camera gain, resulting in increased measurement noise. This noise can potentially impact our optimization process.
In all the rest sequences, Colmap-PCD is affected by varying degrees of errors introduced by feature extraction, matching, and point-to-plane association, resulting in lower performance. Additionally, for those sequences with significant exposure time variance, Colmap-PCD may yield inferior results compared to our method due to the lack of exposure time estimation (e.g. sequence \textit{visual challenge}), or even complete failure due to the difficulty in identifying the proper feature matching (e.g. sequence \textit{hku campus seq 01}).
Furthermore, when evaluating on the R$^3$LIVE dataset, we observed that Colmap-PCD encountered challenges with system initialization, leading to system failures or inaccurate pose estimations. Some qualitative mapping results are illustrated in Fig. \ref{fig:benchmark}, demonstrating the better resolution and consistency of our method in constructing high-fidelity colorized 3D maps. More visualization results can be found in our video at \url{https://youtu.be/jtIUBI0U76c}.

\begin{figure}
    \setlength\abovecaptionskip{-0.2\baselineskip}
    \centering
    \includegraphics[width=0.9\linewidth]{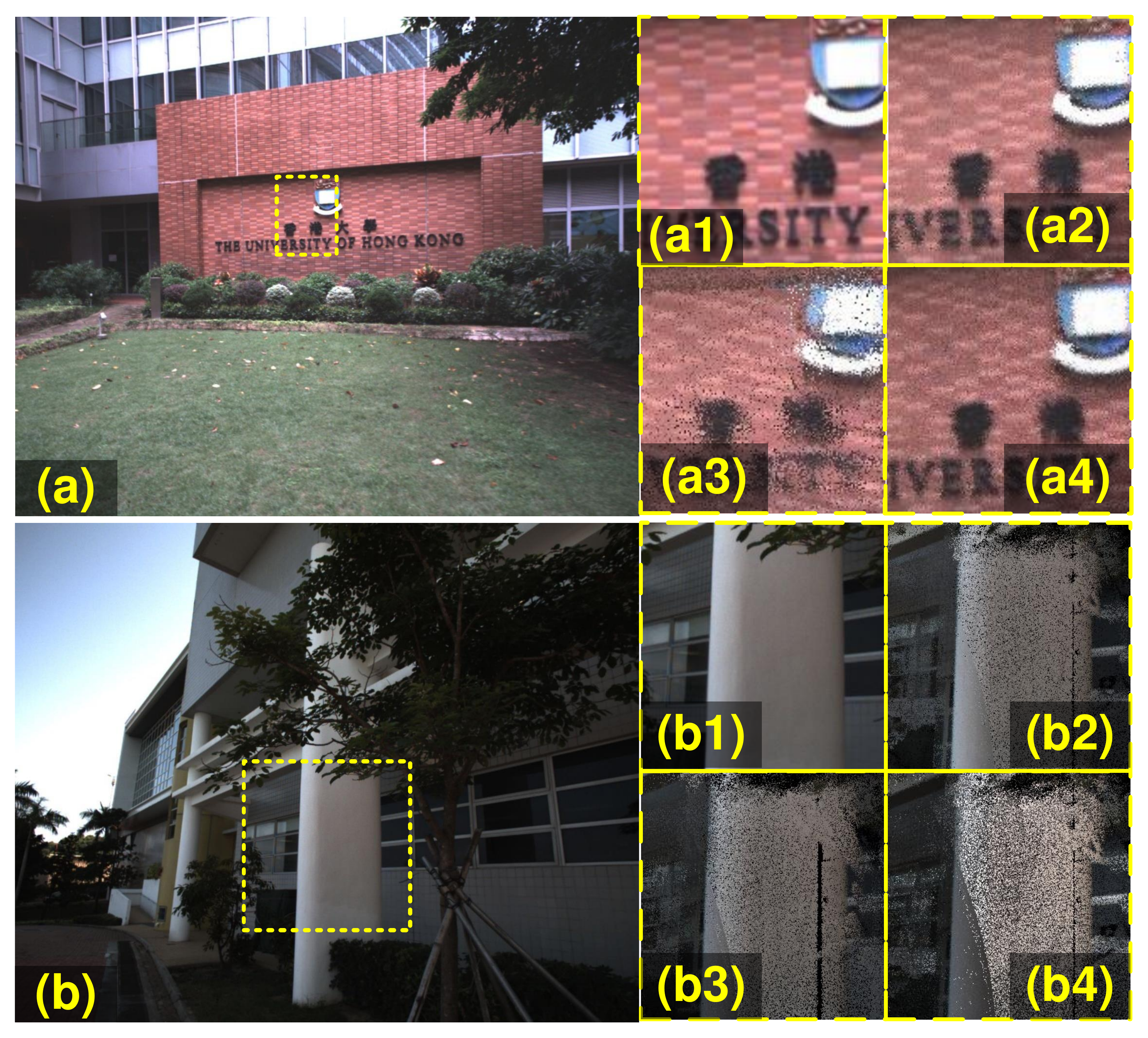}
    \caption{A visualization of our ablation study. (a) and (b) show the original image captured by the camera. (a1) and (b1) show the details of the original image. (a2) and (b2) display the result produced by our full algorithm. (a3) and (b3) depict the outcome when global scene points are removed. (a4) and (b4) represent the result of our method when the estimation of relative exposure times is removed. This sequence effectively demonstrates the impact and contribution of each component in our method to the overall result.}
    \label{fig:ablation_study}
    \vspace{-0.5cm}
\end{figure}

\begin{figure}
    \setlength\abovecaptionskip{-0.2\baselineskip}
    \centering
    \includegraphics[width=1.0\linewidth]{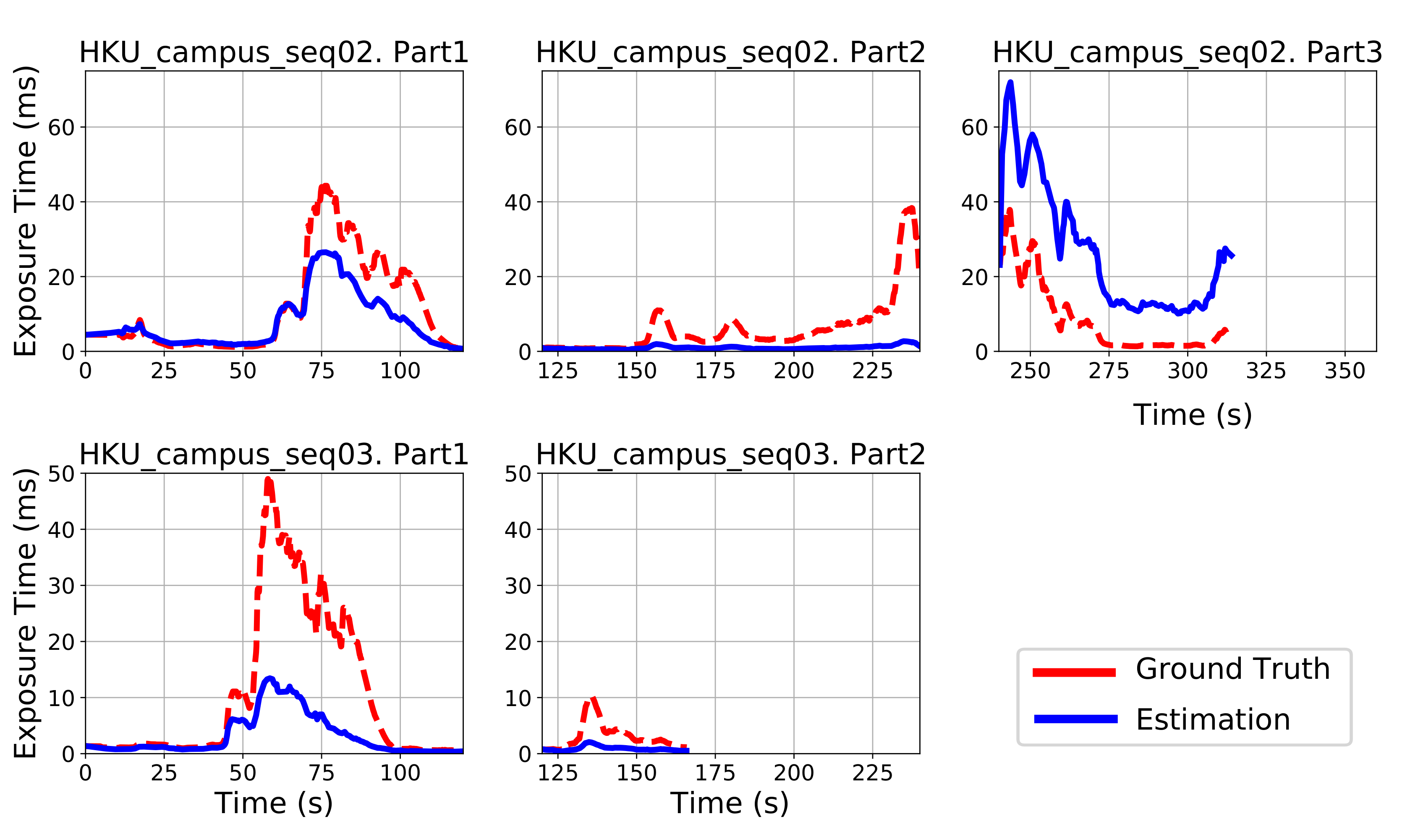}
    \caption{The comparison between our estimated relative exposure time and the ground truth.}
    \label{fig:ret}
    \vspace{-0.5cm}
\end{figure}

Compared with the LVBA without global scene point constraints (``w/o GSP'') or relative exposure time estimation (``w/o RET''), our full algorithm, incorporating global constraints, significantly improves mapping quality. Fig. \ref{fig:ablation_study} illustrates that a lack of global constraints provided by our global scene points results in color blurring and separation in the colorized point cloud map, and failing to estimate relative exposure times leads to noticeable brightness and color discrepancies between the original image and our rendered result. The sequences \textit{hku campus seq 02/03}, which provide the ground truth for camera exposure time, were used to compare our estimated relative exposure times against the actual values. As depicted in Fig. \ref{fig:ret}, while our estimates do not provide the absolute values due to the simplified model, they still successfully capture the general trend of the ground truth exposure times. \label{sec:ret}
\section{Conclusion}
In this paper, we introduced a LiDAR-Visual bundle adjustment framework, aimed at enhancing pose estimation accuracy and ensuring global RGB mapping consistency for LiDAR-camera platforms. By integrating photometric BA for the camera with geometric priors, we achieved high-precision alignment of camera poses with the LiDAR point cloud. Our LiDAR-assisted visibility determination algorithm allowed for the effective global application of this BA method. Through rigorous evaluations, we have demonstrated that our approach surpasses other state-of-the-art methods in both pose estimation accuracy and mapping consistency across multiple datasets. Looking ahead, we aim to incorporate IMU pre-integration and more advanced optimization techniques to further elevate the performance of our system.

\bibliographystyle{IEEEtran}
\bibliography{reference}


\section*{Supplementary material}

# LVBA: LiDAR-Visual Bundle Adjustment for RGB Point Cloud Mapping (Supplementary)

Please note that equation numbers and section numbers from the main manuscript are labeled in this letter in red.

## Appendix A
## Covariance of Simplified Photometric Camera Model

Our method follows [1] and makes some simplification with it. When a ray is projected to a pixel on the camera, the photometric camera model can be represented as

$$\mathbf{I}(\boldsymbol{\rho}) = \mathbf{f}(\tau V(\boldsymbol{\rho})\boldsymbol{\gamma}) \tag{1}$$

where $\boldsymbol{\rho}$ is the pixel position, $\boldsymbol{\gamma}$ is the radiance of the ray, $\tau$ is the camera exposure time, $\mathbf{f}(\cdot)$ is the camera response function (CRF), and $V(\cdot)$ is the vignetting factor accounting for the lens vignetting effect. Since the CRF and vignetting factor of the camera are small but hard to obtain after the sequence is collected, we simply ignore them by setting them to one, leading to a camera photometric model as below

$$\mathbf{I} = \mathbf{f}(\tau V(\boldsymbol{\rho})\boldsymbol{\gamma}) \stackrel{V(\cdot)=1}{\approx} f(\tau\boldsymbol{\gamma}) \stackrel{\mathbf{f}(\mathbf{x})=\mathbf{x}}{=} \tau\boldsymbol{\gamma} \tag{2}$$

Further, we model all sources of measurement noise (e.g. noise from AD transition on camera CMOS) as a Gaussian random noise $\boldsymbol{\delta}_c \sim \mathcal{N}(0, \boldsymbol{\Sigma}_c)$, which will lead to our simplified measurement model:

$$\mathbf{I} = \boldsymbol{\gamma}\tau + \boldsymbol{\delta}_c; \boldsymbol{\delta}_c \sim \mathcal{N}(0, \boldsymbol{\Sigma}_c); \boldsymbol{\Sigma}_c = \begin{bmatrix} \sigma_c^2 & 0 & 0 \\ 0 & \sigma_c^2 & 0 \\ 0 & 0 & \sigma_c^2 \end{bmatrix} \tag{3}$$

where $\mathbf{I}$ is the pixel color we actually measured. $\sigma_c$ is the covariance of measurement noise for each channel.

In LVBA, we assume that all scene points lie on Lambertian surfaces, hence their radiance $\boldsymbol{\gamma}$ is the same in all directions. For two camera frames (i.e. the reference frame and target frames), when their poses are the ground truth and observing the same scene point, the observation radiance should be the same, which indicates:

$$\mathbf{0} = \boldsymbol{\gamma}_r - \boldsymbol{\gamma}_t = \tau_r^{-1}\mathbf{I}_r - \tau_t^{-1}\mathbf{I}_t + (\tau_t^{-1}\boldsymbol{\delta}_{ct} - \tau_r^{-1}\boldsymbol{\delta}_{cr}) \tag{4}$$

where $\boldsymbol{\gamma}_r$ and $\boldsymbol{\gamma}_t$ are the rays projected to the reference and target frames, respectively. $\tau_r$ and $\tau_t$ are the exposure times of the reference and target frames. $\mathbf{I}_r$ and $\mathbf{I}_t$ are the corresponding pixel colors when the rays are captured by the camera. $\boldsymbol{\delta}_{cr}$ and $\boldsymbol{\delta}_{ct}$ are the respective measurement noise. Thus we define the error function $\mathbf{e}$ as:

$$\mathbf{e} := \tau_r^{-1}\mathbf{I}_r - \tau_t^{-1}\mathbf{I}_t \sim \mathcal{N}\left(0, \left(\frac{1}{\tau_t^2} + \frac{1}{\tau_r^2}\right)\boldsymbol{\Sigma}_c\right) \tag{5}$$

The cost item $L$ can be expressed as:

$$L = \|\mathbf{e}\|^2_{(\tau_t^{-2} + \tau_r^{-2})\boldsymbol{\Sigma}_c} \tag{6}$$

The cost function will be the summation of all generated cost items. It should be noted that if we use (6) to construct the cost function and optimize it, the degrees of freedom for the exposure time is not sufficiently constrained. Within a sequence, there is always a general solution: when all exposure time approaches infinity, all the cost items $L$ will tend to zero. To address this problem, we define a "Relative Exposure Time (RET)" and optimize it instead of the exposure time. For the i-th camera frame with the exposure time $\tau_i$, its relative exposure time is defined as:

$$\epsilon_i = \frac{\tau_i}{\tau_1} \quad (i = r, t) \tag{7}$$

where $\tau_1$ is the exposure time of the first camera frame, and there is always $\epsilon_1 = 1$. Then, the error $\mathbf{e}$ could be written as

$$\mathbf{e} = \tau_r^{-1}\mathbf{I}_r - \tau_t^{-1}\mathbf{I}_t = \tau_1^{-1}\left(\epsilon_r^{-1}\mathbf{I}_r - \epsilon_t^{-1}\mathbf{I}_t\right) \tag{8}$$

Since the constant scale factor $\tau_1^{-1}$ only scaled the whole cost function and does not affect the optimal camera state during optimization. We simply ignore it, leading to the error $\mathbf{e}_\epsilon$:

$$\mathbf{e}_\epsilon := \tau_1 \mathbf{e} = \epsilon_r^{-1}\mathbf{I}_r - \epsilon_t^{-1}\mathbf{I}_t$$
$$\sim \mathcal{N}\left(0, \tau_1^2\left(\frac{1}{\epsilon_t^2} + \frac{1}{\epsilon_r^2}\right)\boldsymbol{\Sigma}_c\right) \tag{9}$$

Also, since the value of $\tau_1$ and $\sigma_c$ in our covariance only scaled the whole cost function, which does not affect the optimal camera states during optimization, we simply set $\sigma_c \tau_1 = 1$. And the final covariance matrix $\boldsymbol{\Sigma}$ will be

$$\boldsymbol{\Sigma} = \left(\frac{1}{\epsilon_t^2} + \frac{1}{\epsilon_r^2}\right)\mathbf{E} \tag{10}$$

where $\mathbf{E}$ is the $3 \times 3$ identity matrix. Combining (9) and (10), the final cost item is $\|\mathbf{e}_\epsilon\|^2_{\boldsymbol{\Sigma}}$ as illustrated in (8).

It is worthy mention that implementing the covariance (9) in the cost function $\|\epsilon_t^{-1}\mathbf{I}_t - \epsilon_r^{-1}\mathbf{I}_r\|^2$ is necessary. Otherwise, an apparently wrong optimal solution to the cost function would exist, which is $\epsilon_i = \infty, \forall i$. Incorporating the covariance in the cost function would rectify this issue by excluding the wrong optimum.


\end{document}